\documentclass[letterpaper]{article} 
\usepackage{aaai2026}  
\usepackage{times}  
\usepackage{helvet}  
\usepackage{courier}  
\usepackage[hyphens]{url}  
\usepackage{graphicx} 
\usepackage{natbib}  
\usepackage{caption} 
\usepackage{amsmath,bm,amsfonts}
\usepackage[linesnumbered,ruled,vlined]{algorithm2e}
\usepackage{algorithmic}
\usepackage{multirow}
\usepackage{graphicx}
\usepackage{subfig}
\usepackage{booktabs,makecell}
\usepackage{color}
\usepackage{colortbl} 
\usepackage{xcolor}   

\usepackage{subcaption} 
\usepackage{tabularx}
\usepackage{adjustbox}
\usepackage{hhline}

\usepackage[T1]{fontenc}

\makeatletter
\@addtoreset{subfigure}{figure} 
\makeatother

\definecolor{crimson}{RGB}{220,20,60}
\definecolor{forestgreen}{RGB}{34,139,34}

\usepackage{amsthm}

\usepackage{cite}

\title{Cross-modal Prompting for Balanced Incomplete Multi-modal Emotion Recognition}
\author {
    Wen-Jue~He\textsuperscript{\rm 1},
    Xiaofeng~Zhu\textsuperscript{\rm 2},
    Zheng~Zhang\textsuperscript{\rm 1, \rm 3}\thanks{Corrsponding Author: Zheng Zhang}
}
\affiliations {
    \textsuperscript{\rm 1}Shenzhen Key Laboratory of Visual Object Detection and Recognition, Harbin Institute of Technology, Shenzhen, China\\
    \textsuperscript{\rm 2}School of Computer Science and Technology, Hainan University, Haikou, China\\
    \textsuperscript{\rm 3}Shenzhen Loop Area Institute, Shenzhen, China\\
    he\_wenjue@163.com, seanzhuxf@gmail.com, darrenzz219@gmail.com
}

\usepackage{bibentry}

\begin{document}

\maketitle

\begin{abstract}
    Incomplete multi-modal emotion recognition (IMER) aims at understanding human intentions and sentiments by comprehensively exploring the partially-observed multi-source data.
    Although the multi-modal data is expected to provide more abundant information, the performance gap and modality under-optimization problem hinder effective multi-modal learning in practice, and are exacerbated in the confrontation of the missing data.
    To address this issue, we devise a novel \underline{\textbf{C}}r\underline{\textbf{o}}ss-\underline{\textbf{m}}odal \underline{\textbf{P}}rompting (ComP) method, which emphasizes coherent information by enhancing modality-specific features and improves the overall recognition accuracy by boosting each modality's performance.
    Specifically, a progressive prompt generation module with a dynamic gradient modulator is proposed to produce concise and consistent modality semantic cues.
    Meanwhile, cross-modal knowledge propagation selectively amplifies the consistent information in modality features with the delivered prompts to enhance the discrimination of the modality-specific output.
    Additionally, a coordinator is employed to dynamically re-weight the modality outputs as a complement to the balance strategy to improve the model's efficacy.
    Extensive experiments on $4$ datasets with $7$ SOTA methods under different missing rates validate the effectiveness of our proposed method.

\end{abstract}

\begin{links}
    \link{Code}{https://github.com/WenjueHE/2026-AAAI-ComP}
\end{links}

\section{Introduction}\label{sec:introduction}

The growing demand for human-centric artificial intelligence has stimulated the development of emotion recognition (ER), which is considered as one of the most fundamental tasks in sentiment analysis.
Compared to single-modal emotion recognition, multi-modal emotion recognition (MER) is able to deal with more complicated cases such as satires and metaphors, and has been widely applied to depression detection \cite{fan2024transformer,tao2024depmstat}, dialogue systems \cite{firdaus2020emosen,liang2022emotional}, and so on.

However, the aforementioned traditional MER methods are generally based on an assumption that instances from all modalities are fully observed, which is often broken by limitations in data collection, transmission, and storage in real-world scenarios \cite{Gcnet,he2025dual,zhang2023tensorized}. For instance, background noise might disable the audio modality, and severe accents may lead to failures in speech recognition when constructing the text modality. The incomplete data problem impairs the cross-modal consistency, making it more challenging to achieve accurate emotion recognition.
\begin{figure}[t]
    \centering
    \subfloat[Baseline]{\includegraphics[width=0.48\linewidth]{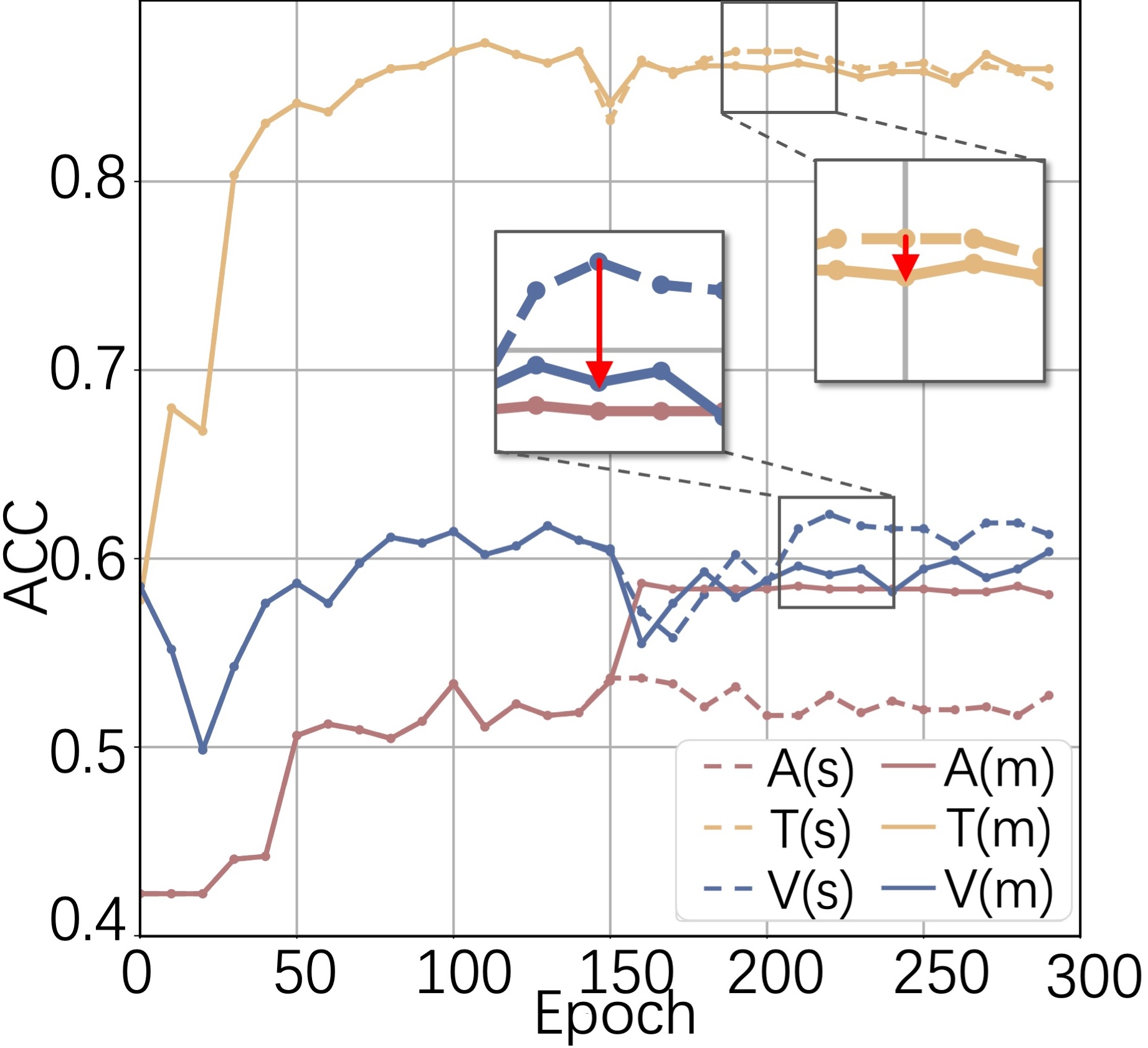}\label{fig:baseline}}
    \hspace{-1mm}
    \subfloat[Ours]{\includegraphics[width=0.48\linewidth]{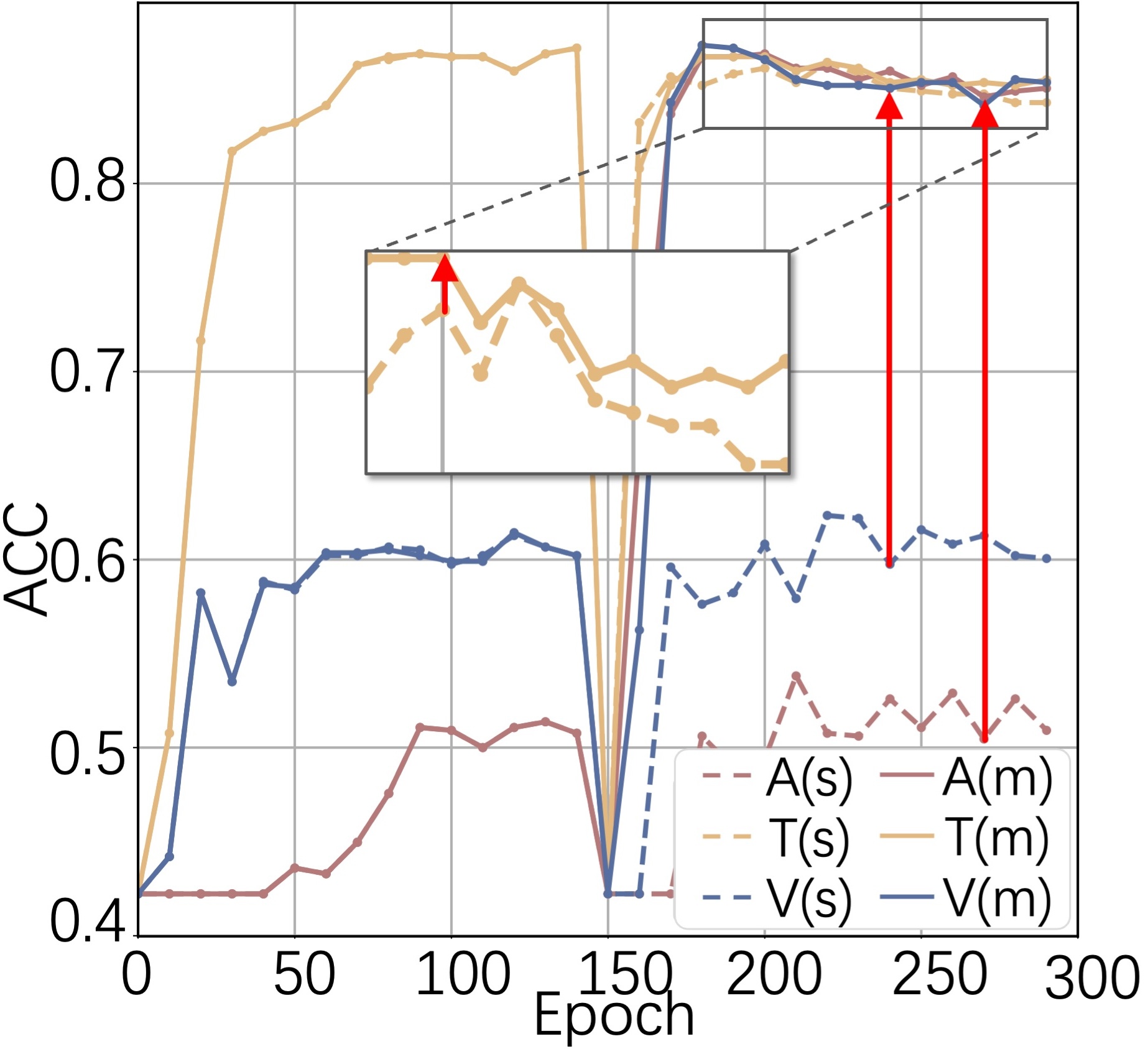}\label{fig:ours}}
    \caption{Modality-specific ACCs of single-modal (s) and multi-modal (m) training of $3$ modalities. \textbf{(a)} In baseline methods, the performance of each modality varies, and the Video (V) and Text (T) modalities suffer a degradation after multi-modal co-training. \textbf{(b)} By cross-modal prompting, all modalities benefit from multi-modal learning.}
    \label{fig:motivation-plot}
\end{figure}

Several attempts have been made to deal with the incomplete MER problem.
Zhao \textit{et al.} (\citeyear{MMIN}) employ the cascaded residual autoencoders (CRAs) to perform conversion from the observed data to the missing ones. 
Taking the data distribution of each modality into consideration, Wang \textit{et al.} (\citeyear{IMDer}) propose to recover the missing instances with the diffusion model by a noising-denoising process. 
Nevertheless, recovering missing data is resource-consuming, and not all the recovered information is necessarily utilized in the training stage. Instead, another stream of methods seeks to learn a unified representation with potential missing modalities. Guo \textit{et al.} (\citeyear{MPMM}) propose different types of prompts to make up for the missing modality information. Li \textit{et al.} (\citeyear{HRLF}) perform multi-grained alignment between the teacher branch pre-trained on the complete data and the student branch trained on the incomplete data. Inspired by the idea of mixture of experts, Xu \textit{et al.} (\citeyear{MoMKE}) feed the modality-specific features to all experts to leverage their knowledge and design a router to determine their weights.

Although progress has been made in IMER, modalities in vanilla multi-modal methods are not fully explored due to the modality imbalance problem \cite{BML}, which could be summarized into modality performance gap, \textit{i.e.}, performances of different modalities vary greatly, and modality under-optimization, \textit{i.e.}, modality-specific performance degrades after multi-modal co-training. Fig. \ref{fig:motivation-plot}(a) presents a showcase of the modality imbalance problem.
Such phenomena could be attributed to the intrinsic heterogeneity of multi-modal data. On one hand, the features of some modalities could be less focused on the task-relevant information than others. On the other hand, the features lie in different spaces, and directly concatenating them together may lead to conflicts. Existing IMER methods generally ignore such discrepancies and treat each modality independently before fusion, therefore fail to address the modality imbalance problem.
Contrary to the existing works, in this paper, we propose to first enhance the representation of each modality by its cross-modal counterparts through knowledge propagation. 
In this way, cross-modal information is delivered to each modality branch to emphasize the emotion-related information that is consistent across modalities, thereby suppressing the misleading information and encouraging the non-dominant modalities to be better utilized.
For more effective consistency enhancement, two questions need to be carefully considered: 

\textbf{Q1: What kind of knowledge should be passed to other modalities? }

\textbf{Q2: How can each modality combine its own information with that of the others?}

\begin{figure*}
    \centering
    \includegraphics[width = \linewidth]{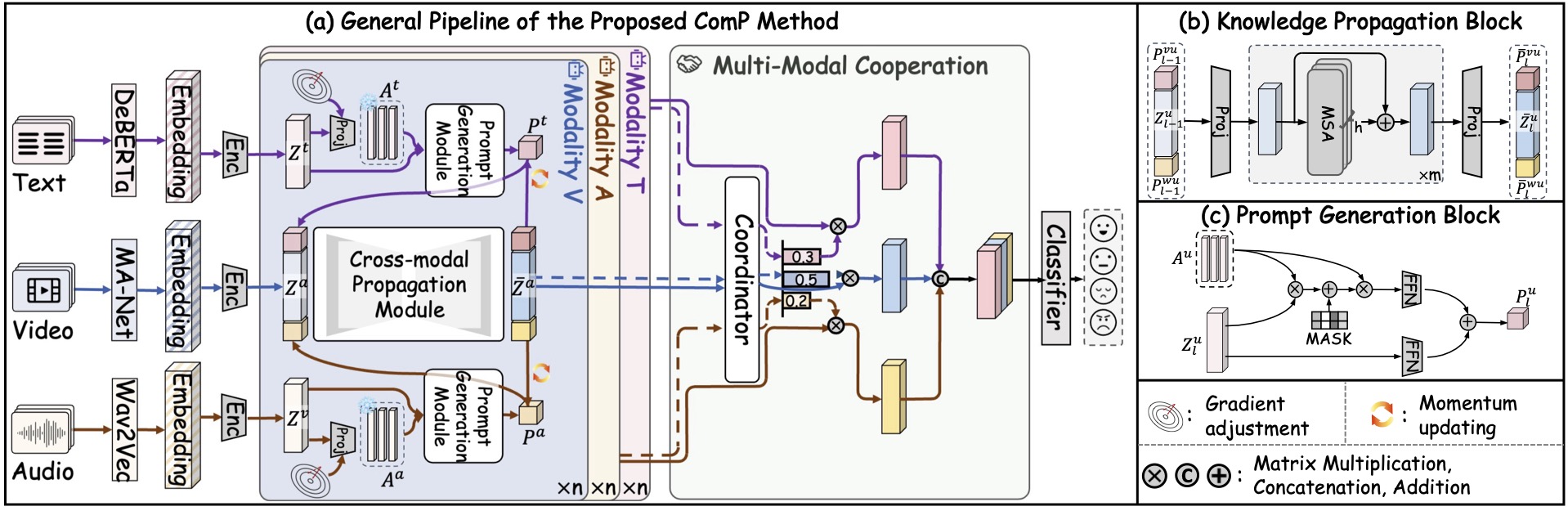}
    \caption{\textbf{(a)} The general framework of the proposed ComP method. First, the prompt generation (PG) module, which consists of several PG blocks \textbf{(c)}, compresses the modality-specific features to obtain consistent yet representative prompts. Then, the prompts are passed to other modalities through the knowledge propagation (KP) module \textbf{(b)} to enhance the task-relevant information. A multi-modal cooperation module (Cr) further examines the significance of each modality and reweighs them before fusion.}
    \label{fig:framework}
\end{figure*}

For the first question, inspired by recent progress in prompt learning \cite{coop,cocoop}, we design a progressive prompt generation module with momentum updating to learn representative and consistent prompts. Specifically speaking, the global input is compressed into a small number of prototypes with gradient modulation to avoid the prototypes being dominated by the easy samples, which are fused with the contextual features to obtain low-dimensional representative prompts. 
For the second question, a simple yet effective cross-modal knowledge propagation module is designed to conduct mutual updating for the modality feature and prompts for invariant information enhancement. Meanwhile, this measure seamlessly reconstructs the incomplete samples without any extra processes. A coordinator is introduced subsequently to rebalance the modality outputs.
Combining the components above, the general framework of our proposed method is illustrated in Fig. \ref{fig:framework}.
As demonstrated in Fig. \ref{fig:motivation-plot}(b), our ComP method well alleviates the modality imbalance problem and boosts the performance of modalities before feeding them into the coordinator.

Our main contributions are listed as follows:
\begin{enumerate}
    \item We propose a novel balanced learning scheme for incomplete multi-modal emotion recognition by promoting each modality to communicate with others. To the best of our knowledge, this is one of the first attempts to alleviate the imbalanced incomplete multi-modal learning problem by prompt learning.
    \item To ensure the quality of broadcasted knowledge, we propose a novel prototype-based progressive prompt generation method with sample-level gradient adjustment, which guarantees the representativeness and consistency of the learned prototypes.
    \item Extensive experiments validate that our proposed method well balances all modalities and reaches superior performance over the existing SOTA methods by promoting each specific modality.
\end{enumerate}

\section{Related Works} \label{sec:related works}
\subsection{Incomplete Multi-modal Emotion Recognition} \label{sec: related-work-IMER}
Compared to early attempts that are based on the assumption of full data accessibility, recent studies have delved into learning from incomplete modalities. 
Some methods try to directly recover the missing instance. 
Deng \textit{et al.} (\citeyear{DENG2025102711}) model the correlations among features by graphs and imputes the missing instances with their k-nearest-neighbors of the rest observed modalities.
Wu \textit{et al.} (\citeyear{DRF}) maintain a feature queue for each modal to simulate the data distribution and conduct cross-modality recovery from both sample and distribution perspectives.
However, directly restoring the original data could be less efficient, as the training process could be resource-consuming, and only a partial restoration of information could contribute to the recognition task. 
Another stream of methods explores the cross-modal connections and learn a consistent representation to mitigate the influence of incompleteness.
Lian \textit{et al.} (\citeyear{Gcnet}) design graph neural networks to simultaneously model the temporal correlation as well as the speaker-wise dependencies.
Inspired by the idea of mixture of experts, MoMKE \cite{MoMKE} feeds the feature of each modality into several experts and designs a router to dynamically adjust their weights.
However, in such methods, the performances of some modalities are generally inferior to others, which results in under-exploration of given data and suboptimal results.

\subsection{Modality-balanced Multi-modal Learning} \label{sec:related-work-balanced-MML}

In multi-modal learning, the model's performance may rely greatly on a single dominant modality \cite{MPT-HCL, LNLN}. In some cases, adding an additional modality has little contribution to the overall performance \cite{vielzeuf2018centralnet}, and may even impair it \cite{wang2020whatmakes}. To address these issues, balanced multi-modal learning methods seek to fully utilize the potential of each modality and reach higher performance.
Structure-based balance learning methods design specific modules to guide the weaker modalities to learn towards the stronger ones. 
Yang \textit{et al.} (\citeyear{yang2024rebalanced}) propose to distill the learnable multi-modal encoders from a pre-trained single modality encoder to learn representative and consistent features for vision-language retrieval.
Fan \textit{et al.} (\citeyear{PMR}) accelerate the learning of slow-learning modalities towards the class prototypes and introduce an entropy regularization to alleviate the suppression from the stronger modalities.
Gradient-based methods, on the other hand, dynamically adjust the gradient for certain variables to ensure sufficient training for the weaker modalities.
BML \cite{BML} suppresses the dominant modality by randomly dropping some of its features as well as decreasing its learning rate to encourage other modalities' training.
Wei \textit{et al.} (\citeyear{MMPareto}) borrow the idea of Pareto efficiency to measure the effect of each modality and avoid gradient conflicts by adjusting the gradient direction.

The abovementioned methods generally focus on the complete modality scenario, and ignore the information loss caused by the unobserved instances that could further exacerbate the modality imbalance problem. On the other hand, our proposed method integrates missing instance management and cross-modality balancing into a unified framework of consensus information enhancing by progressive prompt generation and knowledge propagation, efficiently alleviates the performance gap and modality underoptimization problem in IMER.

\section{The Proposed Method} \label{sec:method}

\subsection{General Framework}
Following existing works \cite{Gcnet,MoMKE}, for each utterance, the audio, text, and video embeddings, denoted as $\bm X^a$, $\bm X^t$, and $\bm X^v$, are first extracted with Wav2Vec \cite{wav2vec}, DeBERTa \cite{DeBERTa}, and MA-Net \cite{MA-Net}, respectively. For clarity, we use $\mathcal{M}=\{a,t,v\}$ to represent all modalities that occur in the experiments, and superscripts $u, v, w \in \mathcal{M}$ to represent certain modalities, which satisfy $u \neq v \neq w$. 

For the incomplete problem, instances in some modalities could be unobserved, which could be suggested by the following indicator:
\begin{equation}
    \gamma_i^u = \left\{
        \begin{aligned}
            1&, \bm x_i^u \in \mathcal{X}_o,\\
            0&, \bm x_i^u \in \mathcal{X}_m. \label{eq:indicator} 
        \end{aligned}        
    \right. 
\end{equation}
In eq. (\ref{eq:indicator}), $\mathcal{X}_o$ and $\mathcal{X}_m$ denote the set consisting of the observed and missing instances, respectively. $\hat{\mathcal{X}} = \{(\gamma^u \textbf{1}_d^{\top}) \odot \bm X^u \vert u \in \mathcal{M}\}$ imputes the missing instances with $0$ for the subsequent processes to eliminate the noise caused by incompleteness, where $\textbf{1}_d^{\top} \in \mathbb{R}^{1\times d}$ is a vector with all elements equals to $1$, $\odot$ represents Hadamard product.

Our goal is to build a function $C:~\hat{\mathcal{X}} \rightarrow \bm Y'$ that maps the incomplete feature $\hat{\mathcal{X}}$ to the emotion predictions $\bm Y'$. 
To achieve this, the model follows a two-stage training scheme. 
In the first stage, a feature updating function $f^u$ and an encoder $\mathrm{Enc}^u$ are jointly trained on each specific modality by minimizing the following objective:
\begin{align}
    \mathcal{L} = \sum_{u\in \mathcal{M}} \mathcal{L}_{\mathrm{enc}}(\bm Z^u, \hat{\bm X}^u) + \sum_{u\in \mathcal{M}} \mathcal{L}_{\mathrm{task}}(\bm Y^u, \bm Y),
\end{align}
where $\mathcal{L}_{\mathrm{enc}}$ and $\mathcal{L}_{\mathrm{task}}$ is the reconstruction and task-relevant loss, respectively, $\bm Z^u = \mathrm{Enc}^u(\hat{\bm X}^u) \in \mathbb{R}^{n\times d}$ is features extracted by  encoders, and $\bm Y^u = \mathrm{Classifier}^u(f^u(\bm Z^u)) \in \mathbb{R}^{n \times 1}$ is the modality-specific predictions.

However, due to the intrinsic characteristics of multi-modal data, the capability of each function $f^u$ to recognize emotions is different. To alleviate such an imbalance and make full use of the multi-modal data, in the second stage, the feature of each modality is compressed to a prompt by the \textit{\textbf{prompt generation module}}, and passed into the \textit{\textbf{cross-modal knowledge propagation module}} of other modalities to enhance the task-relevant consensus information and boost modality performance. Finally, the output of each modality is weighted by a \textit{\textbf{multi-modal cooperation module}} before fusing them together.
A detailed explanation of each module is given in the rest of this section.

\subsection{Cross-modal Knowledge Propagation}\label{sec:method-communication}
To address the modality-imbalance problem that the learning capability of some modalities is inferior to others, we enhance each modality by exposing it to the compressed data from the other modalities. As the emotion signals are consistent across modalities, such a measure could effectively stress the task-relevant information and boost the cross-modal synergistic effect.

Specifically, the knowledge propagation module consists of $n$ corresponding blocks (Fig. \ref{fig:framework}(b)). At the beginning of the $l$-th block in modality $u$, semantical prompts $\bm P^{vu}_l, \bm P^{wu}_l \in \mathbb{R}^{n \times p}$ from modalities $v$ and $w$ (which will be introduced in the next subsection) are passed and concatenated with the block input $\bm Z^u_l$, where we define $\bm Z^u_1 = \bm Z^u$ for clarity.
The concatenated features are first projected to a lower-dimensional fused feature $\bm G^u_{l,0}$:
\begin{align}
    \bm G^u_{l,0} = \mathrm{Linear}_{(d+2p) \rightarrow d}(\mathrm{Concat}([\bm Z^u_l, \bm P^{vu}_l, \bm P^{wu}_l])).\label{eq:propagation-1}
\end{align}
The compressed features are then passed into $m$ multi-head self-attention (MSA) layers with residual connection to further enhance the cross-modal communication. Missing instances are masked in MSA to avoid extra noise:
\begin{align}
    \bm G^u_{l,k} = \bm G^u_{l,k-1} + \mathrm{MSA}(\bm G^u_{l,k-1}), ~1\leq k \leq m,\label{eq:propagation-2}
\end{align}
where $\bm G^u_{l,k} \in \mathbb{R}^{n\times d}$ denotes the output of the $k$-th layer in the $l$-th block. Finally, the compressed features are projected back to the original space to obtain the updated features and prompts:
\begin{align}
    [\bar{\bm Z}_l^u, \bar{\bm P_l}^{vu}, \bar{\bm P_l}^{wu}] = \mathrm{Linear}_{d \rightarrow (d+2p)}(\bm G^u_{l,m}).\label{eq:propagation-3}
\end{align}
$\bar{\bm Z}^u_l \in \mathbb{R}^{n\times d}$, $\bar{\bm P}^{vu}_l \in \mathbb{R}^{n\times p}$, and $\bar{\bm P}^{wu}_l \in \mathbb{R}^{n\times p}$ represents the renewed view-specific feature and prompts passed to modality $u$, respectively. 
For the next block, $\bar{\bm Z}^u_l$ is directly treated as the modality feature, \textit{i.e.}, $\bm Z^u_{l+1}=\bar{\bm Z}^u_l$, while $\bar{\bm P}^{vu}_l$ and $\bar{\bm P}^{wu}_l$ are further processed before usage.
Additionally, after this process, the missing instances are also seamlessly reconstructed by their cross-modal counterparts.

\begin{figure}
    \centering
    \includegraphics[width = 0.95\linewidth]{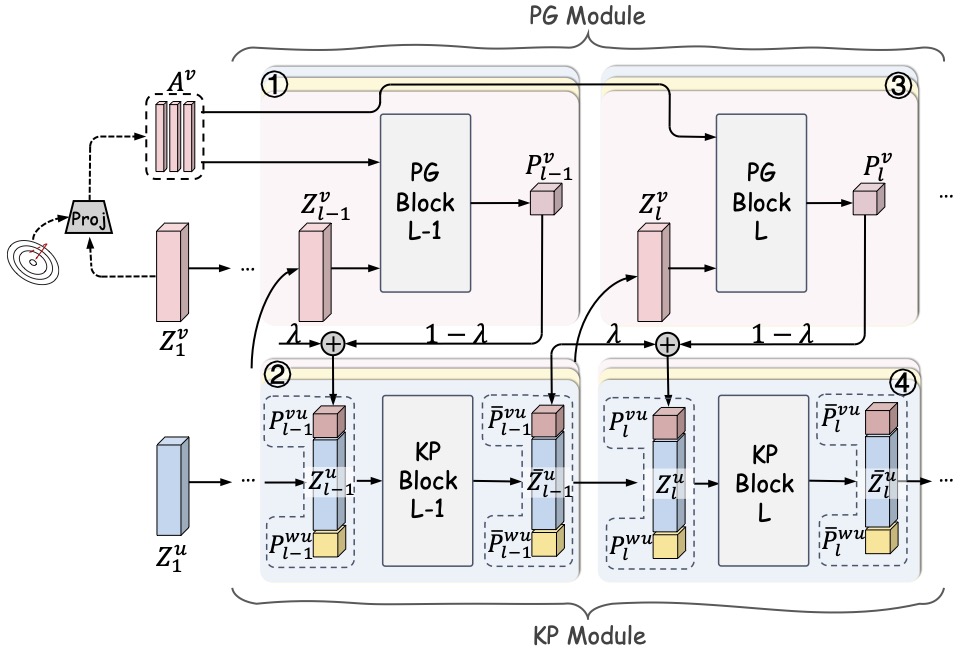}
    \caption{Interactions between the prompt generation (PG) blocks and knowledge propagation (KP) blocks.}
    \label{fig:subframework}
\end{figure}

\begin{table*}[htbp]
    \centering
    \resizebox{\textwidth}{!}{
        \renewcommand{\arraystretch}{1.1}
        \smallskip
        \centering
        \begin{tabular}{ll||ccccccc}
            \Xhline{3\arrayrulewidth}
            \multicolumn{2}{c||}{Missing Rate}                                                       & 0.1                                       & 0.2                                       & 0.3                                       & 0.4                                        & 0.5                                       & 0.6                                       & 0.7                                       \\ \hline
            \multicolumn{1}{l|}{Dataset}                    & Method                               & ACC($\%$)/UA($\%$)                           & ACC($\%$)/UA($\%$)                           & ACC($\%$)/UA($\%$)                           & ACC($\%$)/UA($\%$)                            & ACC($\%$)/UA($\%$)                           & ACC($\%$)/UA($\%$)                           & ACC($\%$)/UA($\%$)                           \\ \hline
            \multicolumn{1}{l|}{}                           & MMIN(ACL '21)                        & 72.30/72.85                                  & 69.22/70.28                                  & 66.21/66.80                                  & 62.77/62.99                                   & 59.89/59.55                                  & 57.87/57.65                                  & 54.23/53.49                                  \\
            \multicolumn{1}{l|}{}                           & GCNet(TPAMI '23)                     & 74.82/74.21                                  & 75.87/75.59                                  & 74.49/74.31                                  & 74.43/73.96                                   & 72.67/72.34                                  & 72.65/73.33                                  & 71.00/70.26                                  \\
            \multicolumn{1}{l|}{}                           & MoMKE(MM '24)                        & 76.70/75.80                                  & 75.25/74.02                                  & 73.47/72.59                                  & 71.42/70.25                                   & 69.73/68.95                                  & 67.95/66.60                                  & 66.52/65.53                                  \\
            \multicolumn{1}{l|}{}                           & SDR-GNN(KBS '25)                     & \underline{78.48/78.25}                                  & \underline{77.83/78.63}                                  & \underline{78.22/78.29}                                  & \underline{76.65/76.32}                                   & \underline{75.47/75.85}                                  & \underline{73.87/74.45}                                  & \underline{70.52/72.21}                                  \\
            \multicolumn{1}{l|}{}                           & \cellcolor[HTML]{EFEFEF}Ours         & \cellcolor[HTML]{EFEFEF}\textbf{80.66/81.09} & \cellcolor[HTML]{EFEFEF}\textbf{79.58/80.22} & \cellcolor[HTML]{EFEFEF}\textbf{78.37/79.16} & \cellcolor[HTML]{EFEFEF}\textbf{77.21/77.96}  & \cellcolor[HTML]{EFEFEF}\textbf{75.62/76.44} & \cellcolor[HTML]{EFEFEF}\textbf{74.28/75.27} & \cellcolor[HTML]{EFEFEF}\textbf{73.41/74.09} \\
            \multicolumn{1}{l|}{\multirow{-6}{*}{IEMOCAPFour}} & \cellcolor[HTML]{EFEFEF}$\Delta$SOTA & \cellcolor[HTML]{EFEFEF}$\uparrow$ 2.18/2.84 & \cellcolor[HTML]{EFEFEF}$\uparrow$ 1.77/1.39 & \cellcolor[HTML]{EFEFEF}$\uparrow$ 0.15/0.87 & \cellcolor[HTML]{EFEFEF}$\uparrow$ 0.56/1.64  & \cellcolor[HTML]{EFEFEF}$\uparrow$ 0.15/0.59 & \cellcolor[HTML]{EFEFEF}$\uparrow$ 0.43/0.08 & \cellcolor[HTML]{EFEFEF}$\uparrow$ 2.41/1.88 \\\hline
            \multicolumn{1}{l|}{}                           & MMIN(ACL '21)                        & 55.21/53.70                                  & 52.00/50.54                                  & 50.25/47.99                                  & 47.51/44.70                                   & 43.79/40.91                                  & 41.41/37.98                                  & 39.47/35.02                                  \\
            \multicolumn{1}{l|}{}                           & GCNet(TPAMI '23)                     & 57.44/57.94                                  & 56.81/54.84                                  & 55.38/53.33                                  & 56.44/54.86                                   & \underline{56.12/54.12}                                  & \underline{54.14/53.62}                                  & \underline{52.59/52.16}                                  \\
            \multicolumn{1}{l|}{}                           & MoMKE(MM '24)                        & \underline{60.54}/58.36                                  & 58.47/56.67                                  & 55.97/53.37                                  & 53.85/52.02                                   & 51.93/49.03                                  & 49.46/47.72                                  & 48.89/46.64                                  \\
            \multicolumn{1}{l|}{}                           & SDR-GNN(KBS '25)                     & 60.26/\underline{59.26}                                  & \underline{58.87/58.78}                                  & \underline{58.80/58.13}                                  & \underline{57.64}/\textbf{56.71}                                   & 53.96/53.07                                  & 53.18/52.01                                  & 51.42/50.29                                  \\
            \multicolumn{1}{l|}{}                           & \cellcolor[HTML]{EFEFEF}Ours         & \cellcolor[HTML]{EFEFEF}\textbf{62.02/61.41} & \cellcolor[HTML]{EFEFEF}\textbf{60.53/58.79} & \cellcolor[HTML]{EFEFEF}\textbf{59.67/58.33} & \cellcolor[HTML]{EFEFEF}\textbf{57.66}/\underline{56.22}           & \cellcolor[HTML]{EFEFEF}\textbf{56.50/54.76} & \cellcolor[HTML]{EFEFEF}\textbf{55.23/53.86} & \cellcolor[HTML]{EFEFEF}\textbf{54.58/53.01} \\
            \multicolumn{1}{l|}{\multirow{-6}{*}{IEMOCAPSix}} & \cellcolor[HTML]{EFEFEF}$\Delta$SOTA & \cellcolor[HTML]{EFEFEF}$\uparrow$ 1.48/2.15 & \cellcolor[HTML]{EFEFEF}$\uparrow$ 1.66/0.01 & \cellcolor[HTML]{EFEFEF}$\uparrow$ 0.87/0.20 & \cellcolor[HTML]{EFEFEF}$\uparrow$ 0.02/$\downarrow$0.49 & \cellcolor[HTML]{EFEFEF}$\uparrow$ 0.38/0.64 & \cellcolor[HTML]{EFEFEF}$\uparrow$ 1.09/0.24 & \cellcolor[HTML]{EFEFEF}$\uparrow$ 1.99/0.85 \\\Xhline{3\arrayrulewidth}
            \end{tabular}
    }
    \caption{Experimental results on IEMOCAPFour and IEMOCAPSix. Best and second best results \textbf{boldfaced} and \underline{underlined}. } \label{table:acc-mr-1}   
\end{table*}

\subsection{Prompt Generation}\label{sec:method-prompt_generation}
Prompts carry the compressed modality information, and play a crucial role in the cross-modality knowledge propagation. 
To effectively highlight the consensus information, the intrinsic emotional information in each modality should be reflected by the prompts, which requires the prompts to be dynamically updated with the features. Additionally, the inter-block consistency should be preserved to avoid extra noise caused by abrupt changes in communication. 
In this section, we present a novel progressive prompt generation method to construct informative and consistent prompts.

\textbf{Original Information Preservation.} 
Learning a small set of prototypes to represent the distribution of the original data has been proven effective and resource-efficient \cite{li2021adaptive, wei2023online}.
Traditional prototype learning methods generally employ a fixed generation scheme such as k-means or averaging, which could be less representative as it cannot be updated.
Contrary to the existing methods, we conduct compression on the batch size dimension and reduce the number of instances to the number of prototypes:
\begin{equation}
    \begin{aligned}
        \bm A^u = (\mathrm{MLP}({\bm Z^u}^{\top}))^{\top}, \label{eq:anchor}
    \end{aligned}
\end{equation}
where $\bm A^u \in \mathbb{R}^{c \times d}$ is the learned prototype matrix with $c$ instances, and each modality-specific Multilayer Perceptron (MLP) is consisted of a $n$-by-$c$ linear layer $g^u({\bm Z^u}^{\top}) = {\bm Z^u}^{\top}\bm M^u  + b^u$, a GeLU activation layer, a $c$-by-$c$ linear layer, and a dropout layer. 
The prototype $\bm A^u$ represents the characteristics of the original feature $\bm Z^u$, and is frozen among all knowledge propagation blocks within each batch.

\textbf{Prompt Updation.} 
As illustrated in Fig. \ref{fig:framework}(c), the obtained prototype $\bm A^u$ is used to represent the global input features $\bm Z^u$. For the $l$-th prompt generation block, given the output of the previous layer $\bar{\bm Z}^u_{l-1}$ and $\bar{\bm P}^u_{l-1}$, the prompts are generated by: 
\begin{align}
    \bm S^u_l &= \mathrm{softmax}(\mathrm{Sim}(\bm Z^u_l, \bm A^u)+\mathrm{Mask}), \label{eq:prompt-1}\\
    \tilde{\bm P}^u_l &=\mathrm{MLP}_p(\mathrm{FFN}_1(\bm S^u_l \bm A^u) + \mathrm{FFN}_2(\bar{\bm Z}^u_l)),\label{eq:prompt-2}
\end{align}
where $\mathrm{Sim}$ denotes the cosine similarity, $\mathrm{Mask} \in \mathbb{R}^{n\times c}$ prevents the missing instances in $\bm Z^u_l$ from disturbing the results by setting the missing instance related entries to a large negative value, $\mathrm{FFN}_1$ and $\mathrm{FFN}_2$ are two feed-forward networks (FFNs), $\mathrm{MLP}_p$ projects the feature to $p$-dim. 
Considering that after the communication process, the prompts passed to other views have also been updated, the final prompts are obtained in a momentum updating manner:
\begin{equation}
    \bm P^{vu}_l = \lambda \bar{\bm P}^{vu}_{l-1} + (1-\lambda) \tilde{\bm P^u_l}, \label{eq:momentum}
\end{equation}
where $0 \leq \lambda \leq 1$ is a hyper-parameter.
In this way, the prompts are updated more stably.

\textbf{Instance-level Gradient Adjustment.} 
Information in some samples could be more conspicuous than the others \cite{AMSS,wei2024enhancing}. 
To avoid the prototypes being dominated by the easy samples, we propose to dynamically quantify the difficulty of each sample by its dependency on cross-modal assistance, and adjust their learning rates to equilibrate the contribution of both simple and challenging samples.

We first calculate the logit error of modality $u$, \textit{i.e.}, $\bm E^u$, which quantifies the distance between the ground truth and the prediction with the index $y$ of the ground truth (gt):
\begin{align}
    \mathrm{logit}^u &= \mathrm{Proj}^u(\bar{\bm Z}^u)_{y}, \nonumber\\
    \bm E^u &= |\mathrm{gt} - \mathrm{logit}^u|.\nonumber
\end{align}

The required modulation of a modality's learning speed is then adjusted based on its relative learning difficulty:
\begin{equation}
    \bm W^u_i = \frac{1}{2} \sum_{v \neq u}  \frac{\sum_{j\neq i} (\bm E^u_j - \bm E^v_j)}{\sum_{j} (\bm E^u_j - \bm E^v_j)}. \label{eq:W}
\end{equation}
$\bm E^u - \bm E^v$ is the relative error between modality $u$ and $v$. 
$\bm W^x \in \mathbb{R}^{n \times 1}$ is multiplied to the gradient of $\bm M^x$ to adjust its learning rate:
\begin{equation}
    \bm M^u_{t+1} = \bm M^u_t - \eta \hat{\bm W}^u \odot \nabla_{\bm M^u}\mathcal{L}(\bm M^u_t),
\end{equation}
where $\hat{\bm W}^x = \bm W^x \textbf{1}_d^{\top} \in \mathbb{R}^{n \times d}$ is the weighting matrix, $\eta$ is the original learning rate, $\odot$ denotes the Hadamard product. In this way, if the relative error of one modality is large, indicating a greater need for other modalities' assistance, its learning rate is decreased correspondingly to give it more chances to pass detailed information to the prototype. Fig. \ref{fig:subframework} shows the details of the interactions between PG and KP blocks.

\subsection{Multi-modal Cooperation} \label{sec:method-coordinator}
After the previously mentioned uni-modal training and cross-modal propagation, the potential of modalities is further explored, and each modality achieves a relatively high performance. 
In this stage, the modalities cooperate with each other to learn a shared representation so that the model can cope with various missing cases. 
Unlike traditional fixed fusing methods such as concatenating or averaging that treat each modality equally, inspired by recent advances in multi-modal fusion \cite{gao2024embracing,MoMKE}, we employ an extra coordinator to dynamically decide how much each modality should contribute to the common representation in different cases:
\begin{align}
    \omega &= [\omega^a, \omega^t, \omega^v] = \mathrm{MLP}(\mathrm{Concat}([\bar{\bm Z}^a, \bar{\bm Z}^t, \bar{\bm Z}^v])),\label{eq:coordinator1}\\
    \bar{\omega}^u &= \mathrm{softmax}(\omega^u) = \frac{\mathrm{exp}(\omega^u)}{\sum_{v \in \mathcal{M} }\mathrm{exp}(\omega^v)}, \label{eq:coordinator2}
\end{align}
where $\bar{\bm Z}^u$ is the output of the KP module.

After obtaining the weighting parameter $\omega_u \in \mathbb{R}^{n\times1}$, the knowledge from each modality is scaled and concatenated, and fed into a classifier to obtain the final prediction:
\begin{align}
    \bm F_i &= \mathrm{Concat}([\bar{\omega}^a_i \bar{\bm z}_i^a, \bar{\omega}^t_i \bar{\bm z}_i^t, \bar{\omega}^v_i \bar{\bm z}_i^v]), \\
    \bm Y' &= \mathrm{Classifier}(\bm F). \label{eq:weighted-concat}
\end{align}
For emotion classification tasks, the model is optimized with the cross-entropy (CE) loss:
\begin{align}
    \mathcal{L}_{\mathrm{task}} = \mathrm{CE}(\bm Y, \bm Y')
                       = -\sum_{i=1}^n \bm Y_i \log \bm Y'_i.
\end{align} 
Otherwise, for emotion recognition tasks, the mean square error (MSE) loss is employed:
\begin{align}
    \mathcal{L}_{\mathrm{task}} = \mathrm{MSE}(\bm Y, \bm Y')
                       = \frac{1}{n}\sum_{i=1}^n (\bm Y_i - \bm Y'_i)^2.
\end{align}


\begin{table*}[htbp]
    \centering
    \resizebox{\textwidth}{!}{
        \renewcommand{\arraystretch}{1.1}
        \centering
        \begin{tabular}{ll||ccccccc}
            \Xhline{3\arrayrulewidth}
            \multicolumn{2}{c||}{Missing Rate}                                                       & 0.1                                       & 0.2                                       & 0.3                                       & 0.4                                        & 0.5                                       & 0.6                                       & 0.7                                       \\ \hline
            \multicolumn{1}{l|}{Dataset}&Method                                                                   & ACC($\%$)/F1($\%$)                           & ACC($\%$)/F1($\%$)                           & ACC($\%$)/F1($\%$)                           & ACC($\%$)/F1($\%$)                            & ACC($\%$)/F1($\%$)                           & ACC($\%$)/F1($\%$)                           & ACC($\%$)/F1($\%$)                           \\ \hline
            \multicolumn{1}{l|}{}                           & MCTN$^\diamondsuit$(AAAI '19)                       & 78.50/78.40                                  & 75.70/75.60                                  & 71.20/71.30                                  & 67.60/68.00                                   & 64.80/65.40                                  & 62.50/63.80                                  & 59.00/61.20                                  \\
            \multicolumn{1}{l|}{}                           & MMIN(ACL '21)                        & 79.88/79.94                                  & 77.74/77.82                                  & 73.78/73.82                                  & 70.43/70.20                                   & 67.07/65.63                                  & 60.98/60.00                                  & 61.43/61.56                                  \\
            \multicolumn{1}{l|}{}                           & GCNet(TPAMI '23)                     & 80.95/81.02                                  & 78.51/78.53                                  & 77.59/77.39                                  & 73.32/73.48                                   & 74.39/74.46                                  & 65.70/65.91                                  & 64.79/64.99                                  \\
            \multicolumn{1}{l|}{}                           & DiCMoR$^\dagger$(ICCV '23)                     & 83.90/83.90                                  & 82.00/82.10                                  & 80.20/80.40                                  & 77.70/77.90                                   & 76.40/76.70                                  & 73.00/73.30                                  & 70.80/71.10                                  \\
            \multicolumn{1}{l|}{}                           & IMDer$^\ddagger$(NIPS '23)                      & 84.90/84.80                                  & 83.50/83.40                                  & \underline{81.20}/81.00                                  & 78.60/78.50                                   & 76.20/75.90                                  & \underline{74.70}/74.00                                  & \underline{71.90/71.20}                                  \\
            \multicolumn{1}{l|}{}                           & MoMKE(MM '24)                        & \underline{86.74/86.74}                                  & \underline{83.38/83.44}                                  & 80.64/80.72                                  & 77.90/78.02                                   & 76.68/76.80                                  & 73.93/\underline{74.08}                                  & 70.58/70.76                                  \\
            \multicolumn{1}{l|}{}                           & SDR-GNN(KBS '25)                     & 82.47/82.32                                  & 81.40/81.19                                  & 80.80/\underline{81.40}                                  & \underline{79.62/79.57}                                   & \underline{78.35/78.26}                                  & 69.98/69.82                                  & 69.32/69.21                                  \\
            \multicolumn{1}{l|}{}                           & \cellcolor[HTML]{EFEFEF}Ours         & \cellcolor[HTML]{EFEFEF}\textbf{87.20/87.11} & \cellcolor[HTML]{EFEFEF}\textbf{85.37/85.23} & \cellcolor[HTML]{EFEFEF}\textbf{83.69/83.67} & \cellcolor[HTML]{EFEFEF}\textbf{81.10/81.18}  & \cellcolor[HTML]{EFEFEF}\textbf{79.27/78.96} & \cellcolor[HTML]{EFEFEF}\textbf{75.61/75.74} & \cellcolor[HTML]{EFEFEF}\textbf{73.17/73.33} \\
            \multicolumn{1}{l|}{\multirow{-9}{*}{CMU-MOSI}}  & \cellcolor[HTML]{EFEFEF}$\Delta$SOTA & \cellcolor[HTML]{EFEFEF}$\uparrow$ 0.46/0.37 & \cellcolor[HTML]{EFEFEF}$\uparrow$ 1.87/1.79 & \cellcolor[HTML]{EFEFEF}$\uparrow$ 2.49/2.27 & \cellcolor[HTML]{EFEFEF}$\uparrow$ 1.48/1.61  & \cellcolor[HTML]{EFEFEF}$\uparrow$ 0.92/0.70 & \cellcolor[HTML]{EFEFEF}$\uparrow$ 0.91/1.66 & \cellcolor[HTML]{EFEFEF}$\uparrow$ 1.27/2.13 \\\hline
            \multicolumn{1}{l|}{}                           & MCTN$^\diamondsuit$(AAAI '19)                       & 81.60/81.80                                  & 78.70/79.00                                  & 76.20/76.90                                  & 74.10/74.30                                   & 72.60/73.60                                  & 71.10/73.20                                  & 70.50/72.70                                  \\
            \multicolumn{1}{l|}{}                           & MMIN(ACL '21)                        & 83.82/83.63                                  & 82.20/81.90                                  & 79.94/79.27                                  & 78.59/77.69                                   & 75.92/75.73                                  & 73.61/71.98                                  & 74.35/72.73                                  \\
            \multicolumn{1}{l|}{}                           & GCNet(TPAMI '23)                     & 85.80/85.82                                  & \underline{85.14/85.10}                                  & \underline{84.48}/84.35                                  & \underline{83.05/82.90}                                   & 82.03/82.08                                  & \underline{81.29}/81.03                                  & \underline{80.05}/80.03                                  \\
            \multicolumn{1}{l|}{}                           & DiCMoR$^\dagger$(ICCV '23)                     & 83.50/83.70                                  & 81.50/81.80                                  & 79.30/79.80                                  & 77.40/78.70                                   & 75.80/77.70                                  & 73.70/76.70                                  & 72.20/75.40                                  \\
            \multicolumn{1}{l|}{}                           & IMDer$^\ddagger$(NIPS '23)                      & 84.80/84.60                                  & 82.70/82.40                                  & 81.30/80.70                                  & 79.30/78.10                                   & 79.00/77.40                                  & 78.00/75.50                                  & 77.30/74.60                                  \\
            \multicolumn{1}{l|}{}                           & MoMKE(MM '24)                        & \underline{85.88/85.65}                                  & 84.76/84.60                                  & 82.61/82.42                                  & 80.99/80.77                                   & 79.09/78.93                                  & 77.49/76.92                                  & 76.00/75.86                                  \\
            \multicolumn{1}{l|}{}                           & SDR-GNN(KBS '25)                     & 85.75/85.63                                  & 85.01/85.06                                  & 84.25/\underline{84.37}                                  & 82.17/82.17                                   & \underline{82.12/82.25}                                  & 80.59/\underline{81.18}                                  & 79.97/\underline{80.54}                                  \\
            \multicolumn{1}{l|}{}                           & \cellcolor[HTML]{EFEFEF}Ours         & \cellcolor[HTML]{EFEFEF}\textbf{86.60/86.48} & \cellcolor[HTML]{EFEFEF}\textbf{85.83/85.59} & \cellcolor[HTML]{EFEFEF}\textbf{85.28/85.10} & \cellcolor[HTML]{EFEFEF}\textbf{84.20/83.94}  & \cellcolor[HTML]{EFEFEF}\textbf{83.30/83.12} & \cellcolor[HTML]{EFEFEF}\textbf{82.17/81.78} & \cellcolor[HTML]{EFEFEF}\textbf{80.82/80.55} \\
            \multicolumn{1}{l|}{\multirow{-9}{*}{CMU-MOSEI}} & \cellcolor[HTML]{EFEFEF}$\Delta$SOTA & \cellcolor[HTML]{EFEFEF}$\uparrow$ 0.72/0.66 & \cellcolor[HTML]{EFEFEF}$\uparrow$ 0.69/0.49 & \cellcolor[HTML]{EFEFEF}$\uparrow$ 0.80/0.73 & \cellcolor[HTML]{EFEFEF}$\uparrow$ 1.15/1.04  & \cellcolor[HTML]{EFEFEF}$\uparrow$ 1.18/0.87 & \cellcolor[HTML]{EFEFEF}$\uparrow$ 0.88/0.60 & \cellcolor[HTML]{EFEFEF}$\uparrow$ 0.77/0.01 \\\Xhline{3\arrayrulewidth}
            \end{tabular}
    }
    \caption{Experimental results on CMU-MOSI and CMU-MOSEI. Best and second best results are \textbf{boldfaced} and \underline{underlined}. Results with $^\dagger$, $^\ddagger$ and $^\diamondsuit$ are from \cite{DiCMoR}, \cite{IMDer}, and \cite{Gcnet}.} \label{table:acc-mr-2}   
\end{table*}

\section{Experiments} \label{sec:experiment}

\subsection{Datasets} 
To comprehensively evaluate the performance of our proposed method, we conduct experiments on four publicly available emotion recognition datasets, \textit{i.e.}, CMU-MOSI \cite{CMUMOSI}, CMU-MOSEI \cite{CMUMOSEI}, IEMOCAPFour \cite{IEMOCAP}, and IEMOCAPSix. 
As all the datasets are originally complete, we follow existing works \cite{MMIN,IMDer} and test them under a synthesized missing condition by randomly removing a proportion of data. The missing rate is defined by $\text{MR} = (\sum_{i=1}^m \mathcal{N}^i)/(mN)$, where $m$, $N$, and $\mathcal{N}^i$ is the number of modalities, total samples, and missing samples in modality $i$. Consistent with \citep{Gcnet}, MR is ranged from $0.1$ to $0.7$ to ensure that each sample is observed in at least one modality.

\subsection{Experiments on Incomplete Data}
The performance of our proposed ComP method, along with $7$ state-of-the-art (SOTA) methods including MCTN \cite{MCTN}, MMIN \cite{MMIN}, GCNet \cite{Gcnet}, DiCMoR \cite{DiCMoR}, IMDer \cite{IMDer}, MoMKE \cite{MoMKE}, and SDR-GNN \cite{SDR-GNN}, is tested in this section.
For a fair comparison, all methods are performed on the same features extracted by pre-trained models as ComP.
For the IEMOCAP dataset, weighted accuracy (ACC) and unweighted accuracy (UA) are adopted to evaluate the performance, while for CMU-MOSI and CMU-MOSEI, ACC and weighted average F1-score (F1) are employed. The test results are shown in Table \ref{table:acc-mr-1} and Table \ref{table:acc-mr-2}.

The recognition accuracies of all methods decline as the MR increases due to the uncertainty and noise introduced by incompleteness.
DiCMoR and IMDer restore the original shallow features with generative models, so that traditional MER methods built on complete datasets could be transferred. 
However, such reconstruction could be noisy and less representative, resulting in relatively inferior performance even at a low MR, with a rapid degradation as MR increases.
GCNet employs an early-fusion strategy by concatenating modality features and passing them through an LSTM, which fails to comprehensively investigate the cross-modality correlation at the utterance level and leads to less accurate recognition.
MoMKE performs cross-modal communication at the model level by passing features to experts pre-trained independently on single modalities. However, such coarse-grained message passing fails to comprehensively exchange feature-level messaging, and its testing results drop severely when the MR becomes large. 

Compared to the existing methods, our knowledge propagation module with prompt learning enables more thorough communication across views by mutual information enhancement, which alleviates the side-effect caused by missing instances and improves the overall performance through single-modal boosting. 
Our ComP almost always outperforms other methods on $4$ datasets with different missing rates. Additionally, the increasing number of missing instances has a relatively small impact on our methods than on others, which again validates that our coherency enhancement strategy could effectively extract the task-relevant information even when some instances are missing.

\begin{table*}[]
    \centering
        \renewcommand{\arraystretch}{1.15}
    
    \resizebox{\textwidth}{!}{
        \begin{tabular}{l|cccc||cccccccccccccc}
            \Xhline{3\arrayrulewidth}
            \multirow{3}{*}{Dataset}     & \multicolumn{4}{c||}{\multirow{2}{*}{Tested Components}} & \multicolumn{14}{c}{Missing Rate}                                                                                                                                                                                                           \\\cline{6-19}
                                         & \multicolumn{4}{c||}{}                                   & \multicolumn{2}{c}{0.1}         & \multicolumn{2}{c}{0.2}         & \multicolumn{2}{c}{0.3}         & \multicolumn{2}{c}{0.4}         & \multicolumn{2}{c}{0.5}         & \multicolumn{2}{c}{0.6}         & \multicolumn{2}{c}{0.7}         \\\cline{2-19}
                                         & KP           & PG          & Cr          & GM          & Acc($\%$)      & F1($\%$)       & Acc($\%$)      & F1($\%$)       & Acc($\%$)      & F1($\%$)       & Acc($\%$)      & F1($\%$)       & Acc($\%$)      & F1($\%$)       & Acc($\%$)      & F1($\%$)       & Acc($\%$)      & F1($\%$)       \\\hline
            \multirow{8}{*}{CMU-MOSI}     & $\circ$      & $\circ$     & $\circ$     & $\circ$     & 83.69          & 83.79          & 81.86          & 81.97          & 78.96          & 79.09          & 75.30          & 75.44          & 72.10          & 72.27          & 69.97          & 70.07          & 68.75          & 68.90          \\
                                         & $\bullet$    & $\circ$     & $\circ$     & $\circ$     & 85.67          & 85.63          & 83.23          & 83.13          & 81.40          & 81.27          & 78.51          & 78.62          & 75.15          & 75.29          & 73.17          & 73.29          & 70.88          & 70.96          \\
                                         & $\bullet$    & $\bullet$   & $\circ$     & $\circ$     & 85.82          & 85.76          & 83.69          & 83.73          & 82.01          & 81.98          & 78.96          & 79.05          & 75.46          & 75.61          & 73.48          & 73.55          & 71.65          & 71.81          \\ 
                                         & $\bullet$    & $\bullet$   & $\bullet$   & $\circ$     & 86.13          & 86.09          & 84.15          & 84.20          & 82.32          & 82.42          & 79.57          & 79.70          & 76.52          & 76.65          & 75.00          & 75.09          & 71.80          & 71.97          \\
                                         & $\circ$      & $\bullet$   & $\bullet$   & $\bullet$   & 84.15          & 84.16          & 81.86          & 81.91          & 79.73          & 79.59          & 77.59          & 77.69          & 73.78          & 73.80          & 71.04          & 71.20          & 69.97          & 69.99          \\
                                         & $\bullet$    & $\circ$     & $\bullet$   & $\bullet$   & 85.82          & 85.73          & 83.69          & 83.66          & 82.32          & 82.28          & 79.42          & 79.48          & 76.52          & 76.66          & 73.32          & 73.44          & 71.19          & 71.09          \\
                                         & $\bullet$    & $\bullet$   & $\circ$     & $\bullet$   & 86.13          & 86.14          & 83.99          & 83.96          & 82.93          & 82.88          & 79.42          & 79.46          & 77.29          & 77.32          & 74.09          & 74.24          & 72.10          & 72.14          \\
                                         & $\bullet$    & $\bullet$   & $\bullet$   & $\bullet$   & \textbf{87.20} & \textbf{87.11} & \textbf{85.37} & \textbf{85.23} & \textbf{83.69} & \textbf{83.67} & \textbf{81.10} & \textbf{81.18} & \textbf{79.27} & \textbf{78.96} & \textbf{75.61} & \textbf{75.74} & \textbf{73.17} & \textbf{73.33} \\\hline
            \multicolumn{5}{l||}{}                                                                  & Acc($\%$)      & UA($\%$)       & Acc($\%$)      & UA($\%$)       & Acc($\%$)      & UA($\%$)       & Acc($\%$)      & UA($\%$)       & Acc($\%$)      & UA($\%$)       & Acc($\%$)      & UA($\%$)       & Acc($\%$)      & UA($\%$)       \\\hline
            \multirow{8}{*}{IEMOCAPFour} & $\circ$      & $\circ$     & $\circ$     & $\circ$     & 72.12          & 71.82          & 70.51          & 69.89          & 68.36          & 67.86          & 65.89          & 65.35          & 62.71          & 62.38          & 59.97          & 59.82          & 57.85          & 57.58          \\
                                         & $\bullet$    & $\circ$     & $\circ$     & $\circ$     & 78.94          & 79.74          & 78.42          & 79.02          & 76.85          & 77.50          & 75.77          & 76.39          & 73.71          & 74.71          & 72.44          & 72.93          & 71.83          & 72.72          \\
                                         & $\bullet$    & $\bullet$   & $\circ$     & $\circ$     & 79.70          & 80.64          & 79.04          & 79.73          & 77.51          & 78.43          & 76.56          & 77.35          & 75.20          & 75.89          & 73.47          & 74.53          & 72.99          & 73.74          \\
                                         & $\bullet$    & $\bullet$   & $\bullet$   & $\circ$     & 80.57          & 81.03          & 79.24          & 80.17          & 78.06          & 78.52          & 76.74          & 77.51          & 75.35          & 76.40          & 73.73          & 74.80          & 73.03          & 74.03          \\
                                         & $\circ$      & $\bullet$   & $\bullet$   & $\bullet$   & 72.09          & 72.16          & 70.60          & 70.18          & 68.13          & 67.24          & 65.49          & 64.76          & 63.08          & 62.32          & 59.57          & 59.59          & 57.36          & 57.05          \\
                                         & $\bullet$    & $\circ$     & $\bullet$   & $\bullet$   & 79.20          & 79.63          & 78.71          & 79.72          & 77.63          & 78.70          & 75.55          & 76.22          & 73.65          & 74.57          & 72.30          & 73.51          & 71.86          & 73.32          \\
                                         & $\bullet$    & $\bullet$   & $\circ$     & $\bullet$   & 79.70          & 80.64          & 78.75          & 79.44          & 78.11          & 78.66          & 76.61          & 77.82          & 75.00          & 75.29          & 73.86          & 74.27          & 72.38          & 72.70          \\
                                         & $\bullet$    & $\bullet$   & $\bullet$   & $\bullet$   & \textbf{80.66} & \textbf{81.09} & \textbf{79.58} & \textbf{80.22} & \textbf{78.37} & \textbf{79.16} & \textbf{77.21} & \textbf{77.96} & \textbf{75.62} & \textbf{76.44} & \textbf{74.28} & \textbf{75.27} & \textbf{73.41} & \textbf{74.09} \\\Xhline{3\arrayrulewidth}
            \end{tabular}
    }
    \caption{Ablation study on 2 datasets. $\circ$ and $\bullet$ denote removed and preserved components, respectively. Best results \textbf{boldfaced}.} \label{table:ablation}
\end{table*}

    


\begin{figure}[]
    \centering

    \subfloat[$\bm Z^v$]{\includegraphics[width=0.25\linewidth]{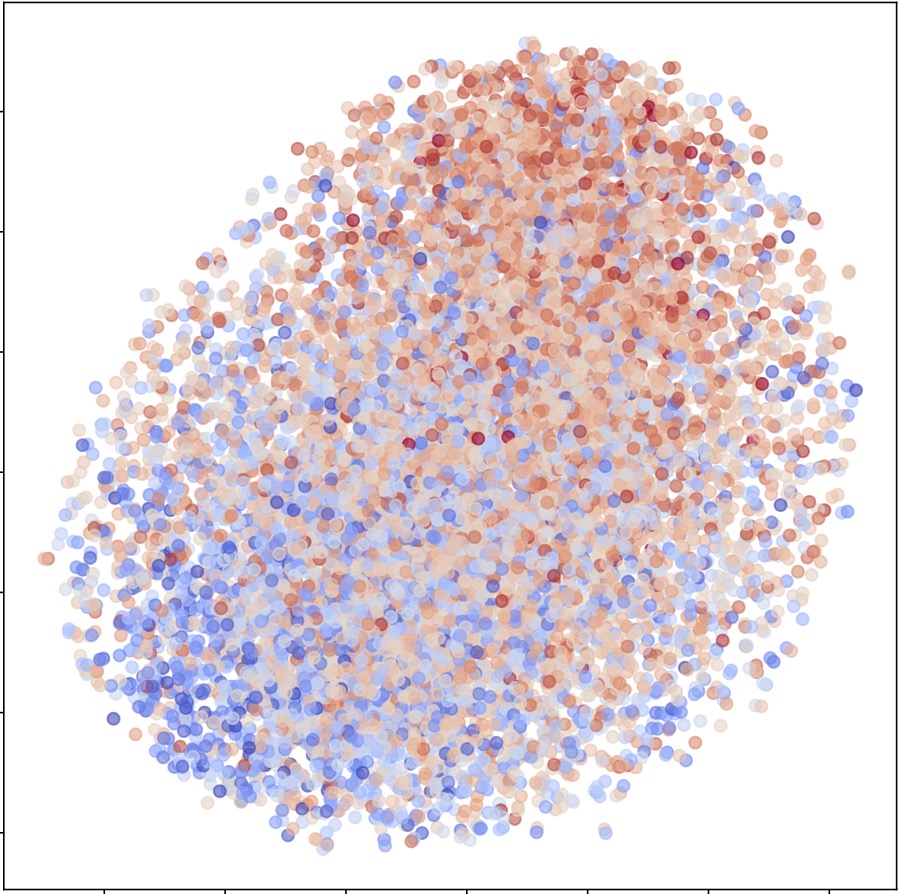}}
    \hspace{2mm}
    \subfloat[$\bar{\bm Z}^v$]{\includegraphics[width=0.25\linewidth]{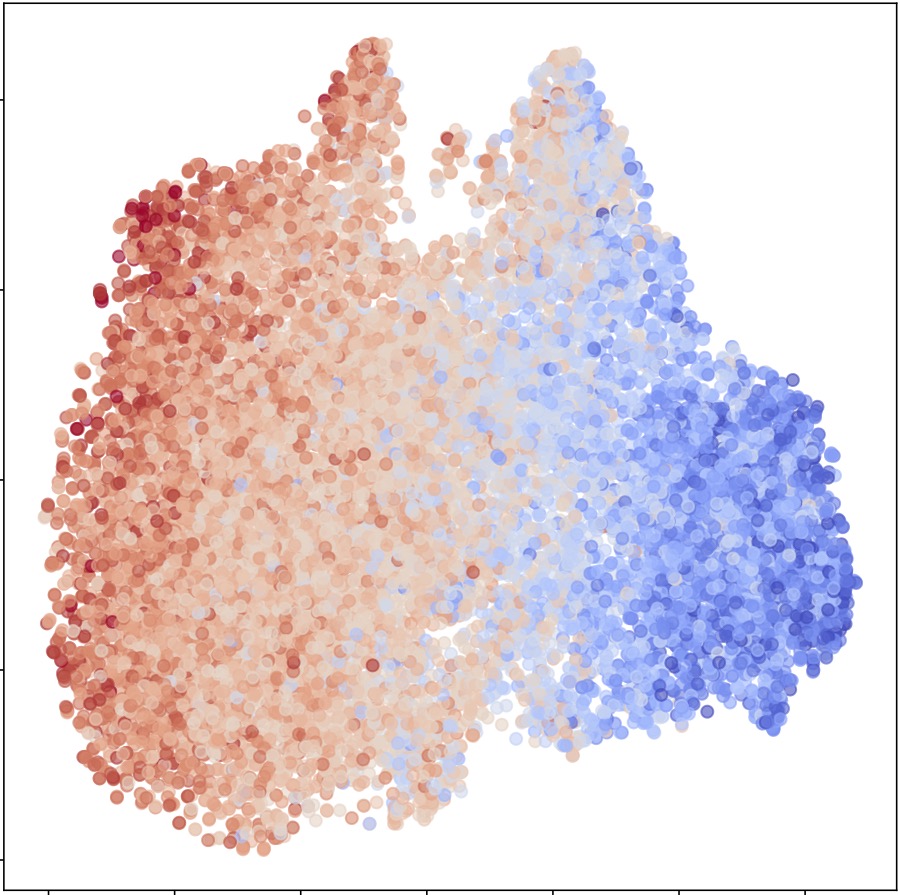}} 
    \hspace{2mm}
    \subfloat[$\bm F$]{\includegraphics[width=0.25\linewidth]{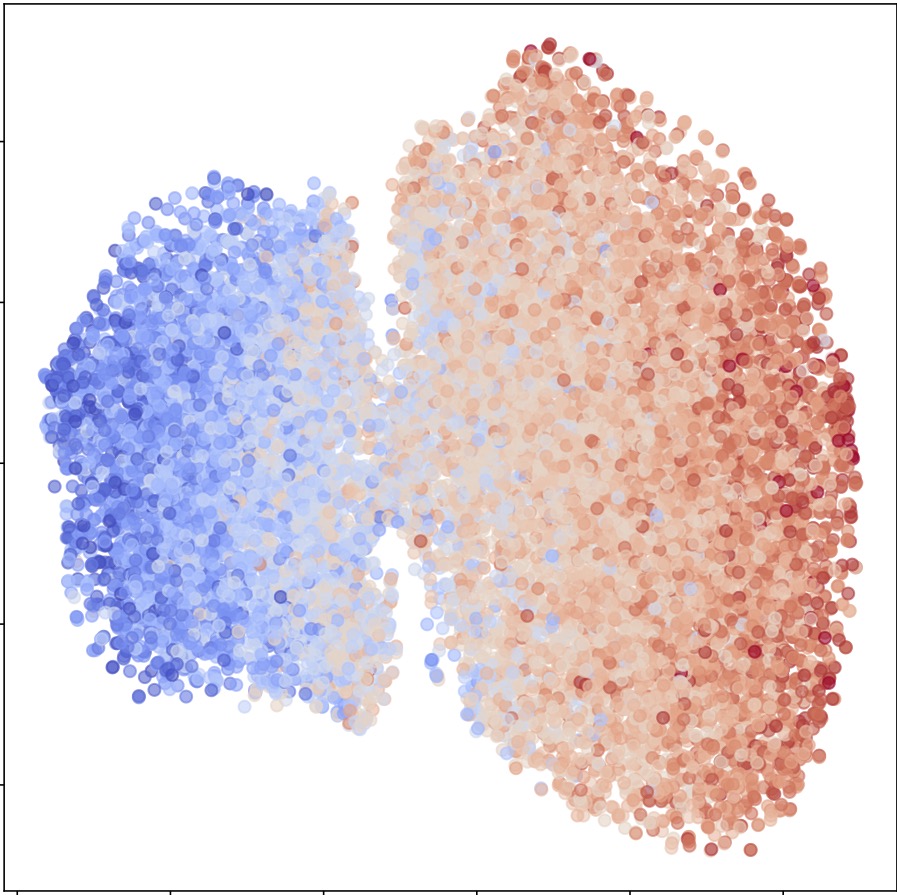}}

    \caption{T-SNE visualization results of $\bm Z^v$, $\bar{\bm Z}^v$, and $\bm F$ on CMU-MOSEI dataset. Positive and negative emotions are marked by red and blue, respectively.}
    \label{fig:tsne-CMUMOSEI}
\end{figure}

\subsection{Ablation Study}

Table \ref{table:ablation} demonstrates the importance of each component. When all components are removed, the output of the encoders is directly concatenated for classification, while when all components are added, the model is equivalent to ComP.
\textbf{1) Knowledge propagation (KP)} is the basis of our proposed method. When KP is removed, the subsequent components are disabled, and the general performance therefore suffers a severe drop, especially when the MR is high. 
\textbf{2) Prompt generation (PG)} aims at generating informative and representative prompts. Removing it adds noise to the propagated knowledge, and also results in a relatively high degradation. 
\textbf{3)} By adding \textbf{modality coordination (Cr)}, the performance of the model continues to increase, indicating that although the modality imbalance problem is alleviated, the employed coordinator could dynamically adjust the importance of each modality.
\textbf{4) Gradient modulation (GM)} further improves the prototype learning in prompt generation, and its promotion effect is more pronounced on CMU-MOSI. The reason could be its various emotion intensity as samples with stronger emotions could easily dominate prototype learning in CMU-MOSI without GM.

\subsection{Visualization}
To further examine the effectiveness of our proposed knowledge propagation strategy, T-SNE visualization of $\bm Z^v$, $\bar{\bm Z}^v$, and $\bm F$, which represent the primary features before propagation, the modality-specific features after propagation, and the fused features after coordination, are illustrated in Fig. \ref{fig:tsne-CMUMOSEI}. 
Different colors represent the emotion spectrum, where saturated red and blue represent the most positive (score $= 3$) and negative (score $=-3$), respectively.
The features before propagation, illustrated in Fig. \ref{fig:tsne-CMUMOSEI}(a), are non-separable. 
After knowledge propagation, the sentiment-related common information is enhanced across all modalities, which can be evidenced by the structured organization of the dots in Fig. \ref{fig:tsne-CMUMOSEI}(b).
Additionally, the overlap between the two parts is further narrowed in Fig. \ref{fig:tsne-CMUMOSEI}(c), demonstrating an enhanced separability of the features achieved by modality cooperation.

\begin{figure}[t]
    \centering
    \subfloat[CMU-MOSEI]{\includegraphics[width=0.48\linewidth]{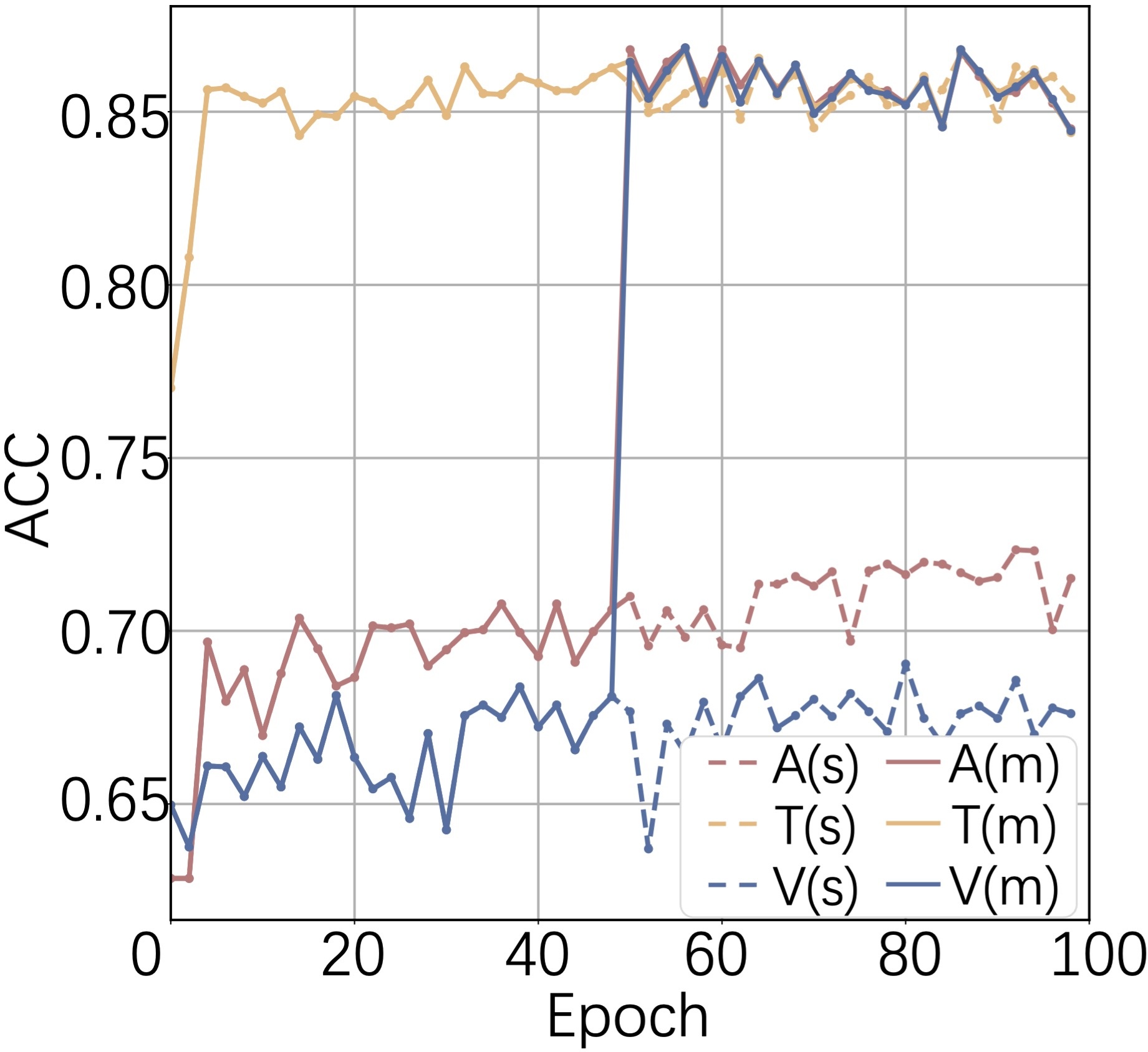}\label{fig:modality-acc-CMUMOSEI}}
    \hspace{2mm}
    \subfloat[IEMOCAPFour]{\includegraphics[width=0.48\linewidth]{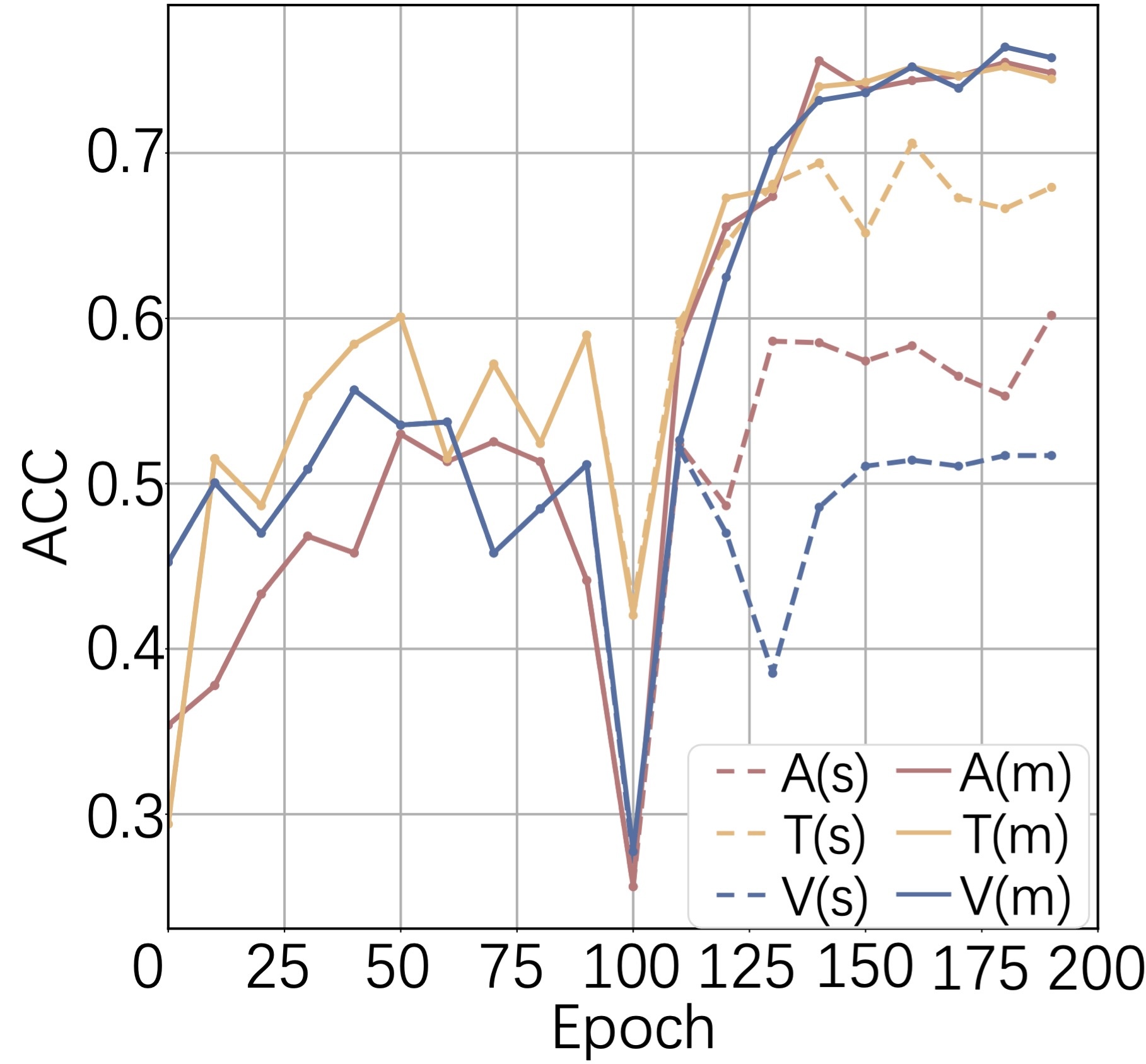}\label{fig:modality-acc-IEMOCAPFour}}\\
    \subfloat[IEMOCAPSix]{\includegraphics[width=0.48\linewidth]{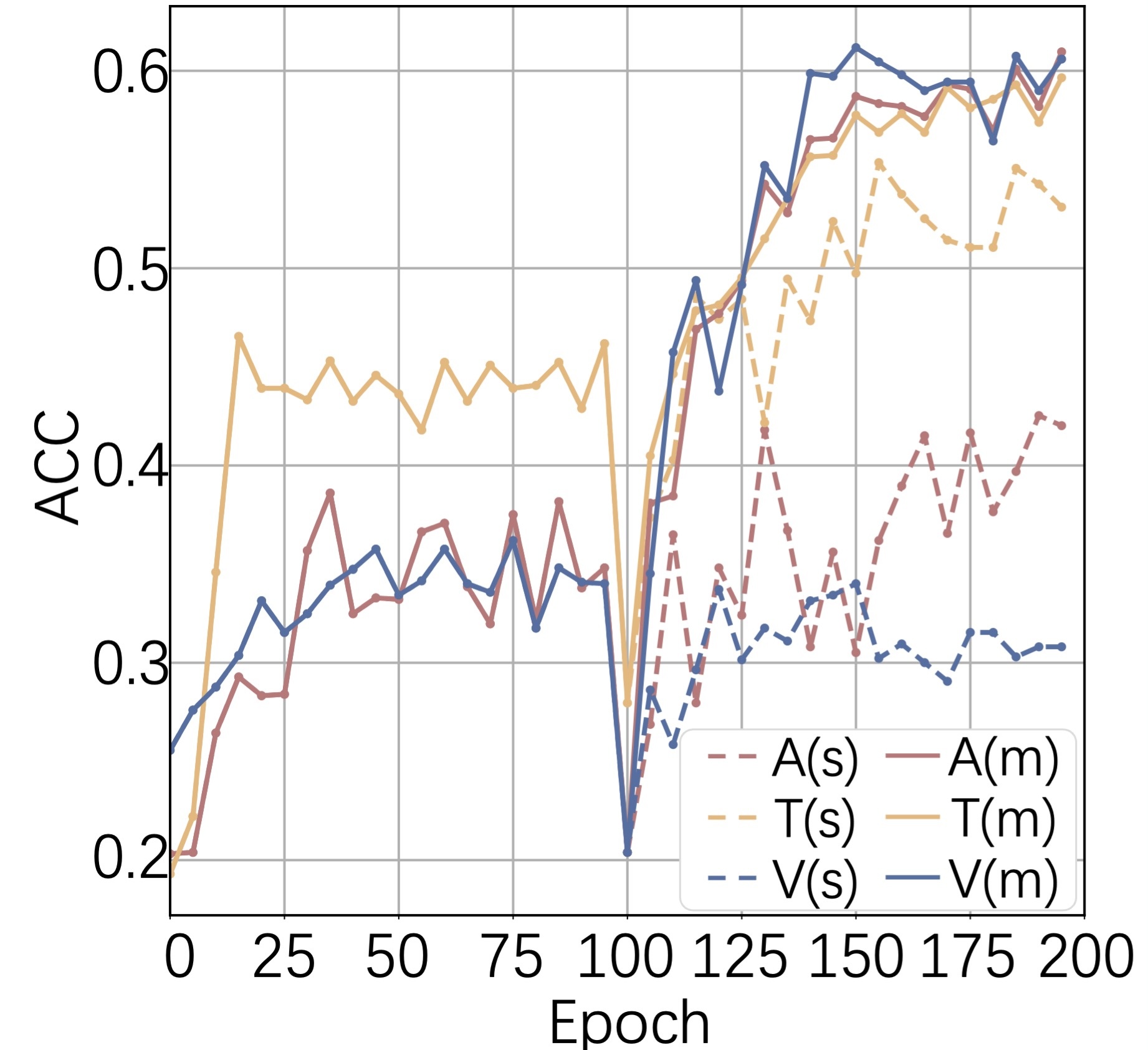}\label{fig:modality-acc-IEMOCAPSix}}
    \hspace{2.3mm}
    \subfloat[Coordinator Weights]{\includegraphics[width=0.48\linewidth]{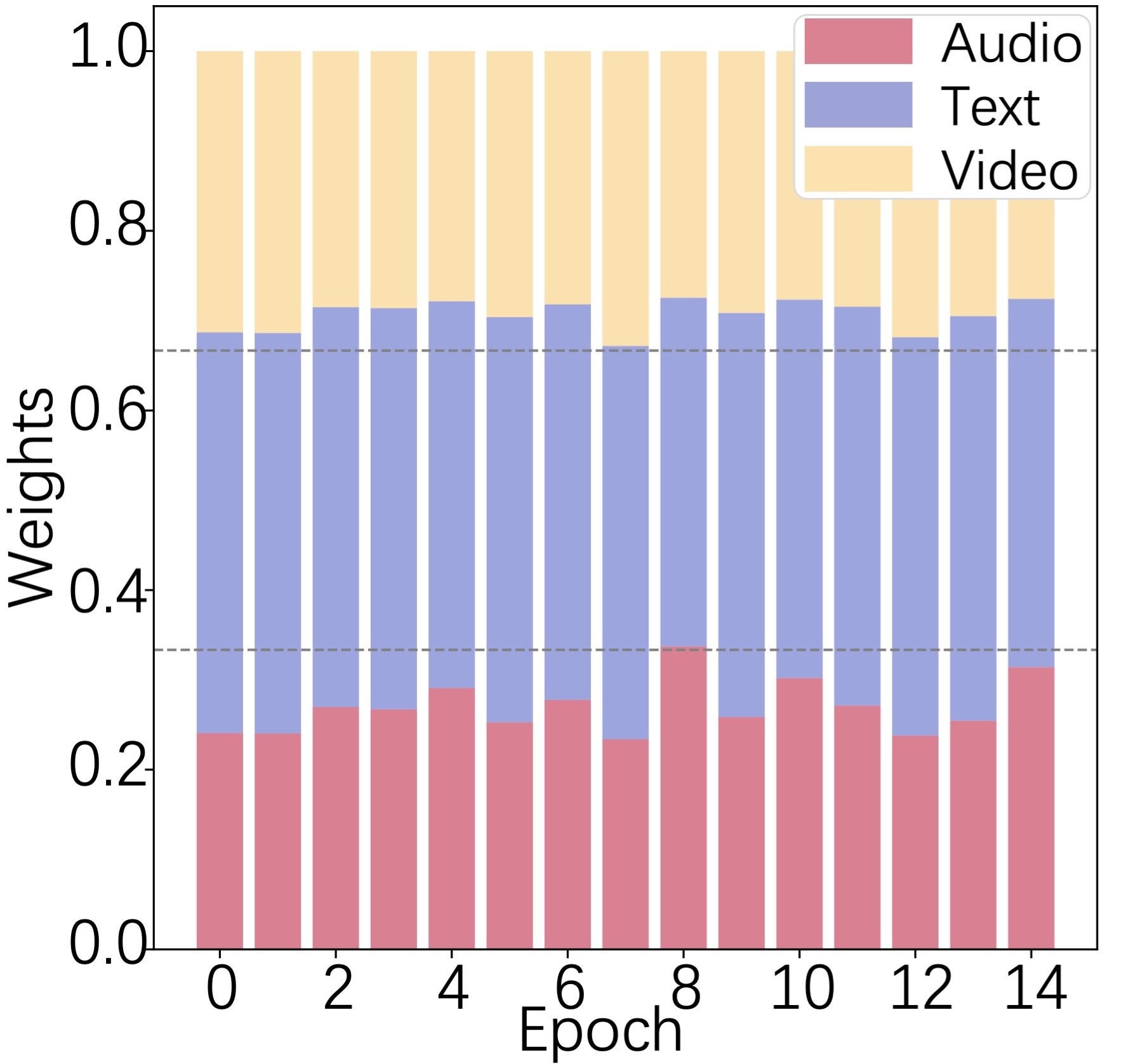}\label{fig:weight}}
    \caption{\textbf{(a-c)} Modality-specific accuracies versus training epochs on $3$ tested datasets. \textbf{(d)} $\bar{\omega}_a$, $\bar{\omega}_t$, and $\bar{\omega}_v$ for different utterances on CMU-MOSI.}
    \label{fig:exp-modality_balance}
\end{figure}

\subsection{Modality-balance Study}
\textbf{Modality-specific Accuracy.}
The modality-specific performances of our ComP method on three datasets are illustrated in Fig. \ref{fig:exp-modality_balance}(a)-(c).
The single modality performances represented in dotted lines vary across all the datasets. Contrarily, the solid lines indicating the single modality performance during co-training experience a sharp improvement at the beginning of the second training stage, where the cross-modal knowledge propagation is involved in the training process. Additionally, for the same modality, the solid line consistently lies above its dotted counterpart, which indicates that our training strategy not only fully explores the potential of each modality, but also jointly utilizes the complementary information from different modalities to reach better emotion recognition performance.

\textbf{Coordinator Weights.}
The weights generated by the coordinator of the utterances from the same conversation are illustrated in Fig. \ref{fig:exp-modality_balance}(d). On one hand, the weights exhibit a relatively balanced distribution across views, 
validating that the discriminative ability of features is comparable across views after knowledge propagation. On the other hand, the weights of the text modality are slightly higher than the other modalities, which showcases the preserved intrinsic of the multi-modal data as well as the necessity of the coordinator.

\section{Conclusion} \label{sec:conclusion}
In this paper, we presented a novel Cross-modal Prompting (ComP) method to address the modality imbalance problems including modality performance gap and modality under-optimization in IMER by task-relevant consensus information enhancement through cross-modality knowledge propagation. Specifically speaking, a global prototype based prompt generation module with gradient regulation was proposed to compress the modality-specific knowledge in a consistent way. Afterward, the consensus information in each modality was enhanced by interactions with cross-modality prompts, so that the accuracy of single modality recognition significantly improves, with the aforementioned modality imbalance problems alleviated. 
Additionally, a coordinator was designed to adaptively adjust the importance of modality fusion, which further ensures the discriminative capability of the proposed method.
Experiments validated that our ComP method not only improves the overall performance, but further explores the potential of each modality.

\section{Acknowledgments}
This research is partially supported by National Natural Science Foundation of China (Grant no. 62372132) and Shenzhen Science and Technology Program (Grant no. RCYX20221008092852077). The authors would also like to thank Huawei Ascend Cloud Ecological Development Project for providing high-performance Ascend 910 processors.

\bibliography{sample-base}

\end{document}